\documentclass[11pt,a4paper]{article}
\usepackage[margin=1in]{geometry}
\usepackage{graphicx}
\usepackage{authblk}
\usepackage{amsmath}
\usepackage{setspace}
\usepackage{listings}
\usepackage{booktabs}
\usepackage{wrapfig}
\usepackage{listings}
\usepackage{dirtree}
\usepackage{xcolor}
\usepackage{fontspec}
\usepackage[hidelinks]{hyperref}
\usepackage{hyperref}
\usepackage{amssymb}
\usepackage{threeparttable}
\usepackage{tikz}
\usetikzlibrary{positioning, calc}
\usepackage{microtype}
\usepackage{array}
\usepackage{makecell}
\hypersetup{
    colorlinks=true,
    citecolor=[RGB]{175,0,0},
    urlcolor=[RGB]{0,60,130},
    linkcolor=[RGB]{0,60,130}
}
\renewcommand{\arraystretch}{1.25}
\definecolor{codebg}{RGB}{250,250,250}
\definecolor{codeframe}{RGB}{220,220,220}
\definecolor{keyword}{RGB}{0,0,120}
\definecolor{importcolor}{RGB}{120,0,0}

\lstdefinestyle{pquant}{
    backgroundcolor=\color{codebg},
    frame=single,
    rulecolor=\color{codeframe},
    basicstyle=\ttfamily\scriptsize,
    keywordstyle=\color{keyword}\bfseries,
    emph={import,from},
    emphstyle=\color{importcolor},
    numbers=left,
    numberstyle=\tiny\color{gray},
    stepnumber=1,
    tabsize=2,
    showstringspaces=false,
    breaklines=true,
    captionpos=b
}
\title{\vspace{-2cm} 
 \includegraphics[width=\textwidth]{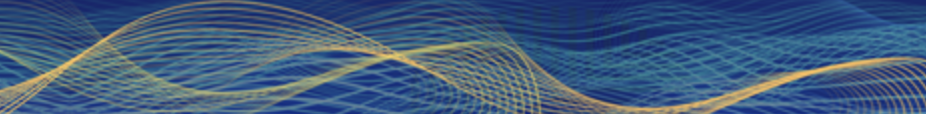} \\[1cm] 
 {\Huge \textbf{PQuantML: A Tool for End-to-End Hardware-Aware Model Compression}} \\[1cm]
 \includegraphics[width=0.25\textwidth]{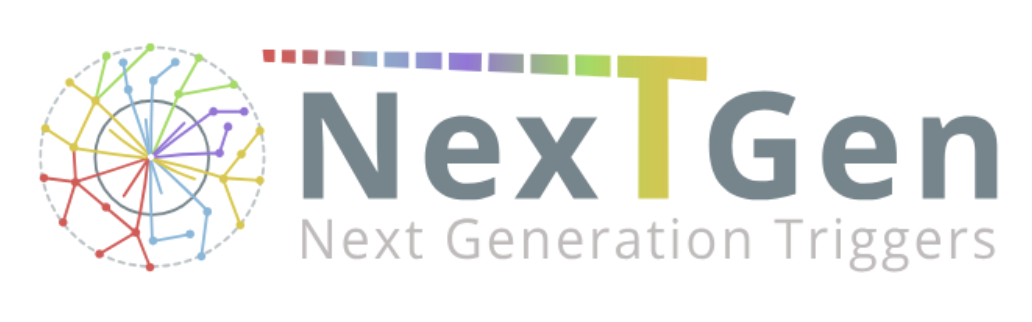} 
}

\author[1,\footnote{Lead contributor and corresponding author: \href{mailto:roope.oskari.niemi@cern.ch}{roope.oskari.niemi@cern.ch}}]{Roope Niemi}
\author[1,\thanks{Primary contributor}]{Anastasiia Petrovych}
\author[2,$\dagger$]{Arghya Ranjan Das}
\author[1,$\dagger$]{Enrico Lupi}
\author[3]{Chang Sun}
\author[1]{Dimitrios Danopoulos}
\author[4]{Marlon Joshua Helbing}
\author[2]{Mia Liu}
\author[5]{Sebastian Dittmeier}
\author[6]{Michael Kagan}
\author[7]{Vladimir Loncar}
\author[1]{Maurizio Pierini}

\affil[1]{European Center for Nuclear Research (CERN), CH-1211 Geneva, Switzerland}
\affil[2]{Purdue University, West Lafayette, IN 47907, USA}
\affil[3]{California Institute of Technology, Pasadena, CA 91125, United States}
\affil[4]{University of Padova, Italy}
\affil[5]{Physikalisches Institut, Heidelberg University, Germany}
\affil[6]{SLAC National Accelerator Laboratory, Menlo Park, USA}
\affil[7]{Institute of Physics Belgrade, Serbia}
\usepackage{xcolor}

\lstdefinestyle{folderstyle}{
  basicstyle=\ttfamily\small,
  frame=single,
  backgroundcolor=\color{gray!5},
  columns=fullflexible,
  keepspaces=true,
  showstringspaces=false
}
\definecolor{kw}{cmyk}{1, 0.50, 0, 0}
\begin{document}
\maketitle
\thispagestyle{empty}

\begin{abstract}
    PQuantML is a new open-source, hardware-aware neural network model compression library tailored to end-to-end workflows. Motivated by the need to deploy performant models to environments with strict latency constraints, PQuantML simplifies training of compressed models by providing a unified interface to apply pruning and quantization, either jointly or individually. The library implements multiple pruning methods with different granularities, as well as fixed-point quantization with support for High-Granularity Quantization. We evaluate PQuantML on representative tasks such as the jet substructure classification, so-called jet tagging, an on-edge problem related to real-time LHC data processing. Using various pruning methods with fixed-point quantization, PQuantML achieves substantial parameter and bit-width reductions while maintaining accuracy. The resulting compression is further compared against existing tools, such as QKeras and HGQ.
\end{abstract}

\newpage
\setcounter{page}{1}

\section{Introduction}

High-energy physics (HEP) experiments at the Large Hadron Collider (LHC) \cite{Evans:2008zzb} face an extreme data challenge \cite{Shiers:2007LHCcomputing}: proton-proton collisions occur at a rate of 40 MHz, producing data at a rate of hundreds of terabytes per second, making it infeasible to store all recorded events offline. To mitigate this, the ATLAS~\cite{atlas-tdr-029} and CMS~\cite{cms-tdr-021} experiments at the LHC employ multi-level trigger systems that rapidly process the collision events and decide which ones to keep for further analysis. The trigger system operates at two levels: a hardware-based Level-1 trigger system (L1T) that makes decisions within microseconds, and a software-based high-level trigger (HLT) that further refines the selection within $\sim$1 second.

Traditionally, the L1T relies on fast but simple algorithms, such as placing thresholds on physics-motivated quantities, while HLT uses more complex domain-knowledge methods~\cite{cms-tdr-021,atlas-tdr-029}. However, many of these advanced algorithms are too slow for direct implementation in the hardware-level trigger. In recent years, machine learning (ML) has emerged as a powerful solution to this problem~\cite{hls4ml2025,qkeras,tgc,Sun_2025,jedilinear,ae-l1t, nn-fpga}, providing fast approximations of computationally expensive algorithms without sacrificing accuracy. ML models can capture complex patterns and, once trained, can be deployed efficiently in real-time pipelines when properly optimized for hardware constraints.

The L1T is implemented on custom electronic boards equipped with field-programmable gate arrays (FPGAs), where algorithms are specifically designed to meet stringent latency and throughput requirements. Due to their ability to perform massively parallel computations with low latency, these devices are also well suited for implementing ML inference tasks.

Efficient deployment of ML models on FPGA hardware requires optimization techniques that reduce resource usage while meeting latency constraints. In particular, pruning removes redundant parameters, while quantization reduces numerical precision. A key approach is to incorporate these techniques directly during training, rather than applying them solely as a post-processing step. Frameworks such as QKeras~\cite{qkeras} enable quantization-aware training, while more recent methods such as HGQ~\cite{hgq} provide finer-grained control over bit-widths and improved hardware awareness. However, these libraries focus primarily on quantization, requiring users to implement pruning strategies separately when more advanced pruning methods are needed.

To address this gap, we introduce PQuantML \footnote{The software is available at: \href{https://github.com/cern-nextgen/PQuantML}{https://github.com/cern-nextgen/PQuantML}.}, a library that integrates pruning and quantization in a unified framework. PQuantML builds upon and extends HGQ quantization features with hardware-aware fine-granularity support, while adding systematic pruning capabilities. Its design emphasizes usability: compression strategies can be specified through configuration files, and the library orchestrates multi-round training and hyperparameter optimization in close integration with user-defined training pipelines. In this way, PQuantML lowers the barrier for physicists to adopt advanced compression techniques in their ML workflows for real-time applications. This paper describes the first stable release of PQuantML and demonstrates its functionality on a benchmark LHC physics task, namely jet tagging.

This paper is structured as follows: Section~\ref{sec:relatedwork} reviews the background of model compression methods and related tools. Section~\ref{sec:design} introduces the design and architecture of PQuantML, while Sections~\ref{sec:workflow} and~\ref{sec:features} detail its workflow and feature set. Section~\ref{sec:benchmarks} presents case studies and benchmarks on representative models for FPGA deployment, demonstrating the effectiveness of pruning and quantization. Sections~\ref{sec:discussion}~and~\ref{sec:future} discuss limitations and future directions, respectively. Finally, Section~\ref{sec:conclusions} concludes with a summary of contributions and outlook for ML deployment in LHC real-time systems.

\section{Background and Related Work}
\label{sec:relatedwork}

Real-time deployment of machine learning models in HEP environments requires methods that drastically reduce computational and memory demands while preserving physics performance. Over the past decade, significant progress has been made in model compression, hardware-aware optimization, and FPGA-focused ML deployment backends~\cite{hls4ml2025, da4ml, finn} as well as frontends. In this section, we review the current landscape of the available frontends and the techniques they use for model compression and FPGA-oriented optimization.

\subsection{Model Compression for Real-Time ML}
Deploying neural networks (NNs) in real-time HEP environments requires a fundamentally different design perspective from offline ML. In offline settings, techniques such as overparameterization are commonly used to improve model generalization~\cite{overparam}. In contrast, such methods are inefficient in real-time systems, where the resulting increase in computational complexity leads to higher latency and excessive on-chip resource usage when implemented on FPGAs. 

In the context of FPGA-based trigger systems at LHC experiments, ML models must satisfy strict latency requirements on the order of microseconds or
less, while simultaneously fitting within a limited hardware budget. FPGA resources, such as digital signal processors (DSPs), block RAM (BRAM), lookup tables (LUTs), and flip-flops (FFs), are limited, and their usage scales directly with the model size, inter-layer dataflow bandwidth, and numerical precision. Moreover, the L1T system must operate with deterministic latency, ensuring every operation completes within a fixed number of clock cycles, regardless of input complexity.

As a consequence, model quality in this domain cannot be evaluated using accuracy or loss metrics alone. Instead, the design space is defined by optimization problems involving at least three objectives: inference latency, hardware resource consumption, and physics performance \cite{aarrestad2021fastcnn}, forming a Pareto frontier, as improvements in one typically require trade-offs in the others. In practice, reaching a desirable operating point requires reducing model complexity without significantly degrading its performance, which can be achieved through careful compression design.

Model compression comprises a broad set of techniques that reduce computational cost by removing redundant parameters, enforcing structurally compact architectures, or applying lower-precision operations. Compression strategies include pruning~\cite{han2016deepcompression,li2017pruning,frankle2019lottery}, quantization~\cite{jacob2018quantization,zhou2016dorefa}, knowledge distillation~\cite{denil2013predicting, pol2023knowledgedistillationanomalydetection}, and architecture-level optimization, such as symbolic regression~\cite{Tsoi_2024}. More recently, low-rank adaptation and factorization methods have also emerged as practical compression tools \cite{lora2}. Pruning and quantization are particularly effective for FPGA deployment~\cite{qkeras,Ramhorst2023FPGA,metamlpro}, where deterministic latency and explicit control over dataflow and numerical precision are critical. Pruning reduces the number of operations and associated memory transfers, while quantization reduces the cost per operation and can improve utilization of FPGA DSP blocks and LUT-based arithmetic ~\cite{jacob2018quantization,zhou2016dorefa,Ramhorst2023FPGA,hls4ml2025}. While post-training pruning and quantization can improve latency and reduce resource consumption, they often induce non-negligible performance degradation at high compression ratios. In contrast, pruning- and quantization-aware training~\cite{Coelho_2021} has been shown to achieve substantially better trade-offs between model size and performance. In particular, successful deployment may require compression-aware training, coordinated sparsity scheduling, accurate simulation of quantized activations, and fine-tuning to recover performance~\cite{qkeras,hls4ml2025,banner2019posttraining}. Without such integration, models that meet latency and resource constraints often fail to satisfy physics performance requirements, or vice versa. Model compression is therefore critical for translating general-purpose neural architectures into designs that comply with the strict latency and hardware constraints of FPGA-based trigger systems. An end-to-end compression methodology is therefore necessary to fully exploit the capabilities of ML in real-time systems planned for future LHC upgrades.

\subsection{Pruning: Structured and Unstructured Approaches}
Pruning aims to reduce the computational footprint of NNs by removing parameters that contribute insignificantly to overall model performance. By eliminating redundant weights or entire computational structures, pruning can significantly reduce inference cost, memory usage and activation bandwidth. These benefits make pruning an important technique for deploying models on resource-constrained hardware, including FPGAs.

Generally, pruning can be classified into three main categories: unstructured,
pattern-based, and structured pruning, each with distinct implications for hardware efficiency.

\textbf{Unstructured pruning} removes individual weights regardless of their location within the tensor, based on criteria such as magnitude \cite{han2016deepcompression} or sparsity-inducing regularization \cite{poptim}. This approach can produce extremely sparse weight matrices while maintaining high accuracy and maximal flexibility \cite{gale2019stateofsparsity}. However, the resulting irregular memory access patterns make it challenging to accelerate on modern hardware, making unstructured sparsity inefficient. For FPGAs, unstructured pruning can yield meaningful reductions in resource utilization, but it becomes more cumbersome in resource-reused designs. These limitations are widely recognized in the pruning literature and hardware-oriented sparsity research \cite{mishra2021acceleratingsparsedeepneural, DBLP:journals/corr/abs-2304-06941}.

\textbf{Structured pruning} removes entire computational units, such as channels, convolutional filters, attention heads, or whole layers, rather than individual weights based on statistics~\cite{li2017pruning, Fang_2023_CVPR}. It can also follow regular sparsity patterns, such as block or cross-shaped pruning in convolutional filters, which do not remove whole units but remain memory- and hardware-efficient. Although structured pruning typically achieves lower sparsity than unstructured methods, it produces highly structured architectures that align with hardware parallelism, enabling real reductions in computational and memory cost. For FPGA-based systems, structured pruning can provide several benefits, such as reduced DSP usage through smaller matrix multiplications and convolution operations, lower activation bandwidth, and simplified routing and control logic~\cite{Ramhorst2023FPGA}.
However, removing entire channels or filters can be destructive when only a subset of weights within those structures is redundant. This can limit achievable sparsity and reduce the flexibility of the compressed model. To address these limitations, semi-structured pruning is used, in particular N:M pruning, where $N$ weights are pruned out of every consecutive group of $M$ weights. This pattern preserves local structure while allowing higher overall sparsity. The regularity of N:M sparsity avoids the irregular memory access patterns of unstructured pruning. For example, NVIDIA's Sparse Tensor Cores exploit a 2:4 structured weight-pruning pattern to double effective matrix multiplication throughput \cite{mishra2021acceleratingsparsedeepneural}. Such semi-structured patterns strike a balance between expressivity and efficiency, enabling higher sparsity than fully structured pruning. Although it achieves lower sparsity than unstructured pruning, it preserves regular structures that are more efficiently utilized by hardware accelerators.

\begin{figure}[t]
    \centering
    \includegraphics[width=\linewidth]{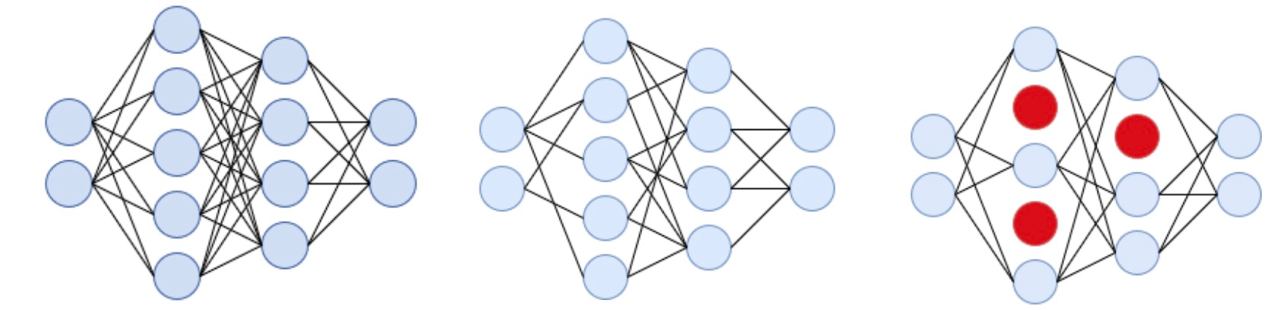}
    \caption{Examples of pruning strategies: original network (left), unstructured pruning with irregular weight removal (middle), and structured pruning removing entire channels or filters (right).}
    \label{fig:pruning_example}
\end{figure}

\subsection{Quantization and Quantization-Aware Training}
Quantization reduces the numerical precision of weights and activations. This work focuses on fixed-point quantization, where weights and activations are represented using low-precision fixed-point numbers rather than common floating-point formats. Lower precision directly decreases memory access and arithmetic cost, which is particularly important for FPGA deployment where resource budgets and latency constraints are strict. In particular, this technique directly reduces DSP, LUT, and FF usage, making quantization a key approach for real-time ML applications in HEP~\cite{jacob2018quantization, qkeras, hgq}.
Quantization can be applied at different stages of the ML workflow:

\textbf{Post-Training Quantization} (PTQ) converts a trained model into lower precision, quantizing the model's parameters, both weights and activations, after training the model. While PTQ is simple and computationally inexpensive, it can suffer from significant accuracy degradation, in particular with aggressive quantization regimes below 8~bits as the model has not been trained to be robust against the quantization noise~\cite{qkeras}. While more advanced quantization schemes, such as channel-wise or weight-only quantization, might mitigate these issues to some extent, they still struggle to recover the model performance when applied to complex architectures \cite{banner2019posttraining}.

\textbf{Quantization-Aware Training} integrates quantization directly into the training loop. In QAT, quantization is applied during the forward pass, and the gradients are propagated through simple approximations such as the straight-through estimator \cite{ste}. In this way, the network learns to operate under quantized constraints and typically retains much higher accuracy at reduced bit-widths. QAT has been successfully used to train 8-bit and down to 4-bit models across a variety of hardware platforms~\cite{jacob2018quantization, qkeras}. QKeras \cite{qkeras} and HGQ \cite{hgq} extend these ideas by supporting fine-grained precision control, per-tensor scaling, and hardware-aware quantization schemes designed to align with FPGA implementation requirements.

\subsection{End-to-end Compression Libraries and Frameworks}
While there are existing software frameworks that support model compression, they often focus exclusively on real-time FPGA deployment in HEP applications. The hls4ml~\cite{hls4ml2025} framework has become the primary toolchain for translating trained networks into firmware using high-level synthesis backends, which provide flexible control over fixed-point precision and parallelism. However, as hls4ml itself does not provide any facilities for pruning or quantization-aware training, the user must rely entirely on external tools to prepare models before conversion. As a result, users must assemble their own isolated quantization and pruning workflows, which can be difficult to maintain.

While quantization is offered in modern machine learning training frameworks, these methods primarily target specialized hardware accelerators, which are not optimal for ultra-low-latency use cases on FPGAs. For instance, PyTorch offers two primary ways to perform quantization: one with fixed-width integer arithmetic and floating point scaling factors, or one using lower-precision floating point formats such as Float16 or Float8. However, since any floating-point operations are prohibitively inefficient, they cannot be directly applied to the target applications considered here.

QKeras~\cite{qkeras} extends the Keras ecosystem by introducing quantized layers and operators designed for low-precision inference and integrates naturally with the hls4ml workflow. It is widely adopted by the HEP community. However, only basic layer-wise and channel-wise fixed-point quantization is supported, limiting the achievable compression ratios and hardware efficiency. Furthermore, the repository has not been actively maintained since 2021, leading to compatibility issues with modern TensorFlow~\cite{tf} or Keras~\cite{keras} versions.

High Granularity Quantization (HGQ)~\cite{hgq} offers more advanced quantization techniques by supporting in-tensor mixed-precision quantizers, per-tensor/rank/value powers-of-two scaling, learned bit-widths through gradients, and hardware-aligned resource consumption estimations. The HGQ framework is tightly integrated with da4ml~\cite{da4ml} and hls4ml workflows for FPGA deployment and it supports JAX~\cite{jax}, TensorFlow, and PyTorch~\cite{torch} backends for training with the help of Keras. In addition to quantization, the framework supports pruning by allowing learned bit-widths to go to zero.

Brevitas \cite{brevitas} enables quantization-aware training in PyTorch and serves the software frontend, tightly integrated with the Xilinx FINN \cite{finn} backend. 

PQuantML offers a unified solution for producing compressed models that meet the strict latency and resource requirements of real-time trigger systems. It integrates structured and unstructured pruning with quantization-aware training and interfaces with hls4ml to translate trained networks into firmware. By combining these features within a user-configurable interface, PQuantML allows users to use and experiment with compression methods.

\section{Library Design and Architecture}
\label{sec:design}

The high-level goal of PQuantML is to provide an accessible, flexible, and hardware-aware framework for model compression and deployment. The library provides an interface for compression-aware training of neural networks prior to deployment and synthesis on hardware such as FPGAs. By emphasizing ease of use and configurability, the library abstracts complex compression workflows into a user-friendly and intuitive interface. At the same time, it maintains hardware awareness, ensuring that compression configurations remain compatible with downstream synthesis and deployment tools. In addition, PQuantML includes an automated hyperparameter optimization pipeline integrated with MLflow \cite{Zaharia2020MLflow} for end-to-end experiment tracking. This integration ensures that checkpoints, configurations, and performance metrics are consistently logged, enabling reproducibility, flexibility, and reusability throughout the model development process.

The overall structure of the PQuantML library is shown in Example~\ref{code:folder_structure} in the Appendix. The library is organized into three main components: the core, data models, and pruning methods.
The \textbf{data models} module defines all configuration objects, which define the functionality for compression, training, and hyperparameter optimization. These configurations are specified via YAML files validated with Pydantic schemas, ensuring both flexibility and type safety. This declarative approach allows users to specify model architectures, pruning and quantization strategies, associated parameters, and experiment metadata in a reproducible manner. Default schemas are provided in the configs directory. \\ The \textbf{pruning methods} module implements the supported pruning algorithms with different strategies and granularities within a unified interface. \\ The \textbf{core} module is divided into PyTorch and Keras submodules, each providing implementations of key building blocks such as \textbf{Quantizer}, \textbf{PQActivation}, compression-aware layers, and the generic training pipeline. The backend is selected at runtime, ensuring compatibility with both PyTorch and TensorFlow. In addition, the core component includes utilities for constructing default configuration objects and managing pruning and quantization parameters during training.
From the user perspective, the import interface remains consistent across backends, providing a unified and framework-agnostic experience. In typical usage, pruning methods do not need to be imported explicitly, as they are configured and applied automatically through PQuantML layers.

\section{Compression Workflow}
\label{sec:workflow}

The compression workflow from model creation to a trained compressed model is managed through a single configuration object that defines corresponding parameters for quantization, pruning, training, hyperparameter optimization, and the FITCompress algorithm~\cite{zandonati2023optimalcompressionjointpruning}. 

Hyperparameter optimization is managed by a dedicated block within the same object, which defines the tunable parameters, their search space, and the number of trials.

\subsection{Model definition}
PQuantML provides compression-aware implementations of standard neural network layers in both Keras and PyTorch, including convolutions, linear layers, pooling, batch normalization, and activations. All activations are handled through a unified \textbf{PQActivation} module, and the framework optionally integrates with external quantization systems such as HGQ or FITCompress. PQuantML supports two complementary workflows for defining compressed models: (i) by constructing layers directly using PQuantML's pruning and quantization modules, and (ii) by applying automatic layer replacement to inject compression operators into an existing architecture.
\subsubsection{Direct Layer}
PQuantML layers are explicitly defined by a user. Data manipulation layers such as \textbf{PQConv2d}, \textbf{PQDense}, \textbf{PQAvgPool2d} function similarly to their native PyTorch or Keras counterparts. As a result, they accept the same constructor arguments with the additional requirement of the configuration object. HGQ is implemented in Keras, and is used through a wrapper class defined in PQuantML. By default, each layer along with its pruning layer and quantizers is initialized using the values defined in the configuration file. However, some hyperparameters, such as enabling quantization, pruning, or specifying custom quantization bits, can be overridden by providing optional arguments when creating the layer. Example~\ref{lst:direct_model} in the Appendix shows a model defined directly using PQuantML layers.

\subsubsection{Layer Replacement}
An alternative way to define the compressed model is to use the layer replacement function. Given the configuration object and a model defined using standard Keras or PyTorch layers, PQuantML replaces each supported layer with a version that includes the appropriate pruning and quantization operators. For finer control, PQuantML supports per-layer configuration through two fields in the configuration object: one controlling pruning and the other defining layer-specific quantization settings, such as bit-widths and whether inputs or outputs are quantized. 

Since model architectures vary, PQuantML provides a helper function that automatically populates both fields with the names of all supported layers. Users can export this template to a YAML file, edit it offline, and use it as the configuration for the construction of compressed models. The layer replacement approach allows users to focus on the model architecture, while the configuration object handles the parametrization of compression layers. It is particularly useful for evaluating pruning methods in isolation across different architectures or for use in the hyperparameter optimization loop of PQuantML (see subsection~\ref{subsec:hypertuning}). 

Example~\ref{lst:layer_replacement} in the Appendix shows how to create a model using the layer replacement function.

\subsection{Model training}
\label{sub:training_func}
\begin{figure}[ht]
    \centering
    \includegraphics[width=0.8\textwidth]{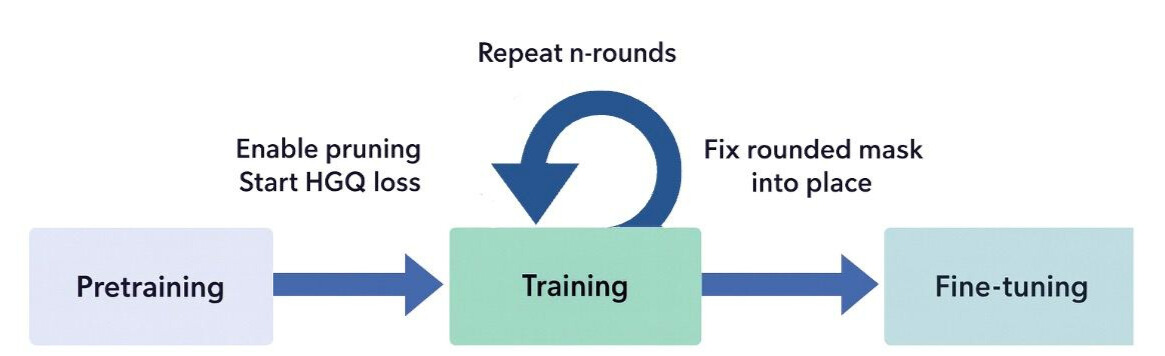}
    \caption{The generic training loop of PQuantML handles the different stages of various compression methods.}
    \label{fig:training_loop}
\end{figure}

Different compression methods follow distinct training sequences: pruning methods range from a single mask-learning stage to up to three stages, while HGQ includes an optional pre-training stage before bit-width optimization. To unify these workflows, PQuantML provides a generic training function (Figure~\ref{fig:training_loop}) controlled entirely by the configuration parameters listed in Table~\ref{tab:training_parameters} in the Appendix.

Training consists of up to three stages: \textbf{pre-training}, \textbf{training}, and \textbf{fine-tuning}. Pruning is disabled during pre-training and enabled during training. Methods that directly learn a hard mask require only the training stage. Methods that learn a soft mask additionally require fine-tuning, during which the soft mask is first rounded to a hard mask and then kept fixed; this stage is also available for hard-mask methods as an optional fine-tuning step with a fixed mask.

For PyTorch models, PQuantML additionally supports FITCompress, which determines per-layer bit-widths and a global sparsity level to meet a target compression ratio. When enabled, quantization is disabled during pre-training; afterwards, the algorithm runs and applies the resulting sparsity and bit-widths to all quantizers and layers before training begins. Example~\ref{lst:train_model} in the Appendix shows a usage example of the generic training function.

\subsection{Converting to FPGA firmware}
Once the model has been fully trained, the final quantization and pruning mask are applied to the weights and biases. The resulting model can then be converted to HLS code for FPGA deployment using the hls4ml package~\cite{hls4ml2025, fastml_hls4ml}. PQuantML is fully compatible with hls4ml, enabling users to follow its standard workflow without additional configuration. The layer precisions are inferred automatically during hls4ml optimization, and input-output quantizers associated with each layer are reconstructed within the internal representation to preserve the original data path. These features ensure that the predictions of the synthesized hls4ml model are bit-accurate with respect to the original PQuantML model, up to floating-point precision.

\section{PQuantML Features}
\label{sec:features}

\subsection{Quantization}
\label{subsec:quantization}
The quantization in PQuantML is handled through \textbf{QAT}, where a network learns during training to minimize the loss caused by quantization. In particular, fixed-point representation is used, parameterized by
three values: \(k\) (whether to include the sign bit, 0 or 1), \(i\) (integer bits, determining the representable range), and \(f\) (fractional
bits, determining the precision). The sum of these bits is the total bit-width of the quantized tensor.
PQuantML supports two quantization modes: (i) tensor-wise, per-channel or per-weight fixed-point quantization, and (ii) High-Granularity Quantization (HGQ).

\textbf{HGQ} learns individual bit-widths for each weight and activation via gradient-based optimization, utilizing a custom loss with two components: an
approximation of the Effective Bit Operations (EBOPs), which balances model accuracy with expected hardware cost,
and an \(L_{1}\)-regularization term that prevents bit-widths from growing excessively. As HGQ learns a
bit-width per weight, it can prune weights when the learned bit-width becomes zero or negative. EBOPs may also be used together with fixed-point quantization to estimate resource usage on FPGA hardware.

Both the inputs and outputs of a layer may be quantized. However, input quantization is generally preferred, as the input bit-width contributes directly to the EBOPs metric. Input and output quantization can be configured globally through the configuration object. All quantization operations are implemented via a wrapper class, \textbf{Quantizer}, for both HGQ and fixed-point quantization. Configurable parameters include default bit-widths for activations and
weights, rounding mode, and overflow handling. The complete list of quantization parameters is summarized in
Table~\ref{tab:quantization_parameters} in the Appendix.

\subsection{Pruning methods}
\label{subsec:pruning}
Pruning methods in PQuantML learn a pruning mask for each layer during the training stage. These methods differ in the
required number of stages and the granularity at which pruning is applied. While some operate at a single-granularity level, others support multiple granularities. The pruning methods are defined via the pruning configuration object and share only two parameters: a global list of layers for which pruning should be disabled, and a global flag that enables or disables pruning entirely.

For models created via the layer-replacement method, the global list of layer names is used to disable pruning on specific layers. For models defined directly with PQuantML layers, pruning is controlled by the global flag, which can be disabled when creating individual layers. Table~\ref{tab:pruning_granularity_stages} summarizes all pruning methods supported in PQuantML, along with their training stages and granularity levels. Each pruning method is summarized below.

\subsubsection{Structured pruning methods}
\noindent\textbf{Activation Pruning} (AP) collects statistics of a layer's outputs and removes neurons whose average activation falls below a user-defined threshold every \(n\) steps. Output activity is defined as being non-zero, thus assuming ReLU activations~\cite{golkar2019continuallearningneuralpruning}.

\vspace{0.2cm}\noindent
\textbf{MDMM}. A pruning method based on the Modified Differential Method of Multipliers~\cite{PlattBarr1988MDMM}, which formulates sparsification as a constrained optimization problem. Instead of relying on heuristic masking or penalty terms, MDMM enforces sparsity through an explicit set of constraint functions applied directly to the model parameters. The optimization problem is defined as $\min_{\theta} \mathcal{L}(\theta) \;\text{subject to}\; C(\theta) = 0$, where $\theta$ denotes the model parameters, $\mathcal{L}$ is the standard training loss, and $C(\theta)$ is a set of functions quantifying the violation of the sparsity constraints. Different constraint metrics $C(\theta)$ allow MDMM to operate at various pruning granularities, including unstructured sparsity targets, hardware-aware pattern-based compression for convolutional layers~\cite{Wang2022PACA} and FPGA-aware resource optimization~\cite{Ramhorst2023FPGA}.

\subsubsection{Unstructured pruning methods}
\noindent\textbf{AutoSparse.}
A ReLU-mask pruning method with a custom backward gradient. The mask is computed as
\[
    \text{sign}(W)\,\times\,\text{ReLU}\bigl(|W| - \sigma(T)\bigr),
\]
where \(T\) is a learnable threshold. The backward gradient is \(1\) for positive values of
\(|W| - \sigma(T)\) and \(\alpha\) for negative values, where \(\alpha\) is decayed over training
~\cite{DBLP:journals/corr/abs-2304-06941}.

\vspace{0.2cm}
\noindent\textbf{Continuous Sparsification} (CS). A variant of Iterative Magnitude Pruning (IMP) \cite{DBLP:conf/icml/FrankleD0C20}. The binary mask is learned by the function \(\sigma(\beta s)\), where \(s\) is a learnable matrix and \(\beta\) is gradually increased during training. When $\beta$ is reset to 1, the positive entries of $s$ are restored to their initial values, while the negative values are preserved. Unlike IMP, which rewinds the weights after every pruning stage, CS rewinds by default only after all rounds are finished, though per-stage rewinding is also supported.

\vspace{0.2cm}
\noindent\textbf{DST}. A ReLU-mask~\cite{NEURIPS2020_83004190} pruning method is similar to AutoSparse, but without the sign and \(\sigma\) operations in the forward pass, and with a modified backward gradient. DST can prune whole layers, so if the pruning ratio of a layer gets too high, it resets the learnable threshold~\cite{DBLP:conf/iclr/LiuXSCS20}.

\vspace{0.2cm}
\noindent\textbf{PDP}. A distribution-based pruning method that captures each layer's weight distribution and computes a threshold consistent with the target sparsity at each epoch~\cite{NEURIPS2023_8f9f4eb3}. A soft mask is created via softmax comparison between weights and the learned threshold value. PDP utilizes a per-layer sparsity budget that is calculated after the pre-training stage. Both N:M and structured variants are available, with the structured variant relying on channel norms rather than individual weight magnitudes~\cite{NEURIPS2023_8f9f4eb3}.

\vspace{0.2cm}
\noindent\textbf{Wanda}. A post-training pruning method that collects activation statistics and computes a pruning score based on input norms and absolute weight values~\cite{DBLP:conf/iclr/Sun0BK24}. When used during
training as a one-shot pruning step, Wanda can use PDP's layer-wise budget estimation. An N:M variant is
also supported. \\
\begin{table}[h!]
    \centering
    \begin{threeparttable}
        \setlength{\tabcolsep}{4.5pt}
        \renewcommand{\arraystretch}{1.15}
        \begin{tabular}{lccccccc}
            \toprule
                         & \textbf{AP} & \textbf{AutoSparse} & \textbf{CS} & \textbf{DST} & \textbf{MDMM} & \textbf{PDP}      & \textbf{Wanda} \\
            \midrule

            \multicolumn{8}{c}{\textbf{Training stages}}                                                                                       \\
            \midrule
            Pre-training &             &                     &             &              &               & \checkmark        & \checkmark     \\
            Training     & \checkmark  & \checkmark          & \checkmark  & \checkmark   & \checkmark    & \checkmark        & \checkmark     \\
            Fine-tuning  &             &                     & \checkmark  &              & \checkmark    & \checkmark        &                \\
            \midrule

            \multicolumn{8}{c}{\textbf{Granularity}}                                                                                           \\
            \midrule
            Unstructured &             & \checkmark          & \checkmark  & \checkmark   & \checkmark    & \checkmark        & \checkmark     \\
            N:M          &             &                     &             &              &               & \checkmark $^{*}$ & \checkmark     \\
            Structured   & \checkmark  &                     &             &              & \checkmark    & \checkmark        &                \\
            \bottomrule
        \end{tabular}
        \caption{Training stages and pruning granularities for supported pruning methods.}
        \label{tab:pruning_granularity_stages}
        \begin{tablenotes}
            \footnotesize
            \item[*] not yet implemented in PQuantML v0.0.5.
        \end{tablenotes}
    \end{threeparttable}
\end{table}

\subsection{Hyperparameter Tuning}
\label{subsec:hypertuning}
Hyperparameter selection is crucial for obtaining optimal performance from NN models. To identify the best configurations, a hyperparameter optimization loop is employed that repeatedly trains the same model under different hyperparameter settings and evaluates their performance according to a defined objective function. By executing these experiments in parallel, the framework explores the hyperparameter search space by sampling diverse configurations and comparing their resulting metrics.

To automate this process, Optuna \cite{optuna} is used, a state-of-the-art framework for hyperparameter optimization based on Bayesian search, adaptive sampling, and dynamic trial management. These samplers enable precise control over parameter domains, supporting uniform, log-uniform, and step-based constraints. Each trial uses user-defined training and validation functions, and Optuna evaluates the returned objective metric, such as validation loss, accuracy, or any custom-defined score, to assess the performance of the tested configuration. The framework also supports multi-objective optimization, enabling the simultaneous optimization of competing criteria, for instance, maximizing model accuracy while minimizing inference latency or FLOPs, and producing a Pareto front of optimal trade-off solutions.

To ensure reproducibility and systematic experiment tracking, MLflow \cite{mlflow2018} is used to log configurations, metrics, and model artifacts generated during tuning. From a user's perspective, a configuration file specifies the search spaces, parameter ranges, and samplers, while the API interface defines the training, validation, and objective functions alongside the optimization direction. This design provides a unified and extensible interface that allows users to integrate hyperparameter optimization features directly into their existing codebase.

\section{Results on Benchmark Models}
\label{sec:benchmarks}

\subsection{Experimental setup}
We evaluate PQuantML on the jet substructure classification (JSC) task using two datasets from hls4ml. The first dataset contains 16 high-level features (HLF) and was originally published on Zenodo \cite{pierini2019hls4ml}. It is available in two versions, hosted on CERNBox~\cite{cernbox-jet} and OpenML~\cite{openml-jet}.

The second dataset~\cite{jet_dataset} provides particle-level features (PLF) as two-dimensional representations of individual particles, which are flattened before being fed into the network. The JSC task is a five-class classification problem, where the model must identify the particle initiating the jet.

For both versions of the HLF dataset (input dimension 16), a fully connected network is used with three hidden layers of 64, 32, and 32 units, followed by a 5-unit output layer. For the PLF dataset, the input dimension depends on the maximum number of particles $N \in \{8, 16, 32, 64, 128\}$ and the number of features per particle $n \in \{3, 16\}$, where events with fewer than $N$ particles are zero-padded. In our evaluations 128 particles are used with 3 features, giving an input dimension of 384. The corresponding model consists of four hidden layers with 64 units each and a 5-unit output layer. All models are trained with a batch size of 1024 using the Adam optimizer with a learning rate of $1 \cdot 10^{-3}$.

For both tasks, a bit-width of 8 is used for all weights and activations, with 16-bit input and output quantization. The weight bit-width remains fixed per layer, while integer and fractional bits may be defined per-tensor ($^t$), per-channel ($^c$), or per-weight ($^w$). In per-tensor quantization, integer and fractional bits remain fixed during training. In contrast, for per-channel and per-weight quantization, the integer bit is computed during the forward pass to match the dynamic range of the weights, and the fractional bit occupies the remaining bit-width (taking sign bit into account). For the PLF JSC results, the resource usage is reported for both da4ml and hls4ml to facilitate comparison with HGQ results. With da4ml, the RTL design is generated using its functional API, where the forward pass is defined using NumPy-like syntax.

\begin{table}[ht!]
    \centering
    \setlength{\tabcolsep}{3.15pt}
    \renewcommand{\arraystretch}{1.2}
       \footnotesize
    \resizebox{\textwidth}{!}{%
    \begin{tabular}{lccccccc}
        \toprule
        \textbf{Model}                        & \textbf{Acc.} & \textbf{DSP} & \textbf{LUT} & \textbf{FF} & \textbf{Latency[cycles]} & \textbf{F$_{max}$(MHz)} & \textbf{II [cycles]} \\
        \midrule

        \multicolumn{7}{l}{\textbf{HLF JSC (OpenML)}}                                                                                                                                 \\
        PQuantML (no pruning)                  & 76.9\%        & 1175         & 60,089       & 21,201      & 11(54.6ns)               & 201.3                   & 2                    \\
        PQuantML (DST$^{t}$)                  & 76.3\%        & 147          & 6,298        & 4,034       & 10(42.3ns)               & 236.3                   & 1                    \\
        PQuantML (DST$^{w}$)                  & 76.3\%        & 154          & 3,895        & 2,899       & 10(47.2ns)               & 211.8                   & 1                    \\
        PQuantML (PDP$^{t}$)                  & 75.89\%       & 90           & 5,540        & 4,464       & 12(47.1ns)               & 254.8                   & 1                    \\
        PQuantML (PDP$^{c}$)                  & 76.2\%        & 205          & 4,605        & 3,991       & 11(46.5ns)               & 236.8                   & 1                    \\
        PQuantML (FITCompress$^{c}$)          & 76.0\%        & 45           & 7,904        & 3,226       & 8(36.2ns)                & 221.1                   & 1                    \\

        HGQ~\cite{da4ml}                      & 76.3\%        & 15           & 5,209        & 799         & 3(14ns)                  & 215                     & 1                    \\
        QKeras (HLS)~\cite{Ramhorst2023FPGA}* & 76.3\%        & 175          & 5,504        & 3,036       & 15(105ns)                    & > 142.9                 & 2                    \\

        \midrule
        \multicolumn{7}{l}{\textbf{HLF JSC (CERN Box)}}                                                                                                                               \\
        PQuantML (PDP$^{t}$)                  & 74.55\%       & 139          & 6,111        & 4,511       & 10(47.0ns)               & 212.9                   & 1                    \\
        PQuantML (DST$^{t}$)                  & 74.8\%        & 155          & 5,862        & 4,159       & 10(42.1ns)               & 237.5                   & 1                    \\
        PQuantML (DST$^{w}$)                  & 74.8\%        & 142          & 3,515        & 2,953       & 10(45.0ns)                & 222.2                   & 1                    \\
        QKeras (NMI'21)~\cite{qkeras}**         & 72.3\%        & 66           & 9,149        & 1,781       & 11(55ns)                     & ~200                    & 1                    \\

        \midrule
        \multicolumn{7}{l}{\textbf{PLF JSC (128,3)}}                                                                                                                                  \\
        PQuantML (DST$^{t}$)                  & 76.7\%        & 1025         & 36k          & 19.3k       & 14(70.3ns)               & 199                     & 1                    \\
        PQuantML (DST$^{t}$)***                & 76.7\%        & 0            & 44k          & 22.2k       & 12(55.7ns)               & 215.3                   & 1                    \\
        HGQ ~\cite{Sun_2025}***        & 76.7\%        & 0            & 101k         & 19k         & 7(35ns)                  & N/A                     & 1                    \\

        \bottomrule
    \end{tabular}}

    \caption{Resource usage, latency, and model accuracy across PQuantML and existing hardware-aware compression frameworks. In all PQuantML results, the bit-widths of weights and activations stay the same during training, but integer bits and fractional bits of weights are shared per-tensor ($^t$), per-channel ($^c$) or per-weight ($^w$). The  F$_{max}$ of the model marked with * is cited from \cite{hgq}. For the model marked with ** the F$_{max}$ is based on the HLS target clock period. hls4ml is used in all the results except for those marked with ***, where da4ml is used.}
    \label{tab:resources_12816}
\end{table}

\subsection{Performance analysis and discussion}
Table~\ref{tab:resources_12816} reports performance and resource usage after place-and-route using Vitis HLS 2023.2 with a target clock period of 5~ns on a Xilinx Virtex UltraScale+ VU13P FPGA (xcvu13p-flga2577-2-e).

The DST pruning method performs well even under per-tensor quantization. In contrast, per-weight quantization with fixed bit-width significantly reduces LUT and FF usage, but introduces a slight increase in latency. On the OpenML dataset, achieving accuracy above 76\% with PDP and per-tensor quantization proves challenging, with substantial degradation observed beyond 85\% sparsity. Switching to per-channel quantization improves accuracy, although at the cost of increased DSP usage.

FITCompress produces models with mixed weight bit-widths (5 and 6). After applying per-channel quantization and fine-tuning, this approach results in a small accuracy drop compared to other methods, while reducing DSP usage relative to PDP and DST. On the OpenML dataset, DST matches the accuracy of QKeras (76.3\%), while PDP shows a slight decrease (75.9\%). However, both methods significantly reduce resource usage and latency. For example, DST reduces LUT usage from 5,504 to 3,895 and latency from 105\,ns to approximately 47\,ns. FITCompress further reduces DSP usage (down to 45) and improves latency (36.2\,ns), at the expense of a minor increase in LUT usage and a small drop in accuracy (to 76.0\%).

Finally, compared to HGQ, all pruning methods achieve similar accuracy, but incur higher resource usage and latency, showing that while HGQ remains more hardware-efficient, PQuantML provides competitive accuracy with flexible trade-offs between resource usage and performance.

A detailed description of the hyperparameters is provided in Table~\ref{tab:training_hyperparams} in the Appendix.
\subsection{Comparison of HGQ performance}
Since PQuantML also supports HGQ, models trained with PQuantML using HGQ are evaluated against models trained directly with the HGQ library to assess whether both approaches yield similar accuracy and resource usage. On the OpenML dataset, three models are trained using identical hyperparameter settings and the PyTorch-backend for Keras layers. The $\beta$ parameter, which scales the EBOPs term in the HGQ loss, is varied across the three models to obtain different compression levels. Each model is trained for 5000 epochs with a learning rate of $5 \cdot 10^{-3}$ and a batch size of 33\,200. The final trained models are then used to estimate resource usage and evaluate accuracy.

Table~\ref{tab:pquantml_hgq} shows that both approaches produce comparable results across all three configurations. Minor differences in accuracy and resource usage are observed, but these remain within the expected variation between training runs.

\begin{table}[h!]
    \centering
    \setlength{\tabcolsep}{3.15pt}
    \renewcommand{\arraystretch}{1.2}
   \footnotesize
    \resizebox{\textwidth}{!}{%
    \begin{tabular}{lccccccccc}
        \toprule
        \textbf{Model} & \textbf{$\beta$} & \textbf{Acc.} & \textbf{DSP} & \textbf{LUT} & \textbf{FF} & \textbf{Latency[cycles]} & \textbf{F$_{max}$} & \textbf{II [cycles]} \\
        \midrule
        PQuantML + HGQ & $1 \cdot 10^{-6}$            & 76.5\%        & 33           & 10,966       & 4,016       & 9(40.7ns)                 & 220                & 1                    \\
        HGQ            & $1 \cdot 10^{-6}$            & 76.5\%        & 28           & 11,372       & 4,906       & 9(39.5ns)                 & 228                & 1                    \\
        \midrule
        PQuantML + HGQ & $3 \cdot 10^{-6}$             & 76.0\%        & 15           & 5,339        & 2,244       & 9(36.9ns)                 & 244                & 1                    \\
        HGQ            & $3 \cdot 10^{-6}$            & 75.9\%        & 23           & 5,435        & 2,444       & 9(38.2ns)                 & 235                & 1                    \\
        \midrule
        PQuantML + HGQ & $5 \cdot 10^{-6}$             & 75.6\%        & 16           & 4,169        & 1,738       & 8(28.3ns)                 & 283                & 1                    \\
        HGQ            & $5 \cdot 10^{-6}$             & 75.7\%        & 10           & 3,975        & 1,603       & 8(29.6ns)                 & 270                & 1                    \\
        \bottomrule
    \end{tabular}}

    \caption{Comparison of HGQ-based models trained with PQuantML and native HGQ. The models were trained using the same hyperparameter settings with varying $\beta$ values. The resource usage is reported after place-and-route.} 
    \label{tab:pquantml_hgq}
\end{table}

Overall, PQuantML achieves strong performance under high compression, outperforming QKeras when combining pruning and fixed bit-width quantization. In addition, PQuantML's HGQ-based models match the performance of models trained directly with HGQ.
\section{Discussion}
\label{sec:discussion}
PQuantML provides a unified platform for model pruning, quantization, and hyperparameter optimization, enabling an automated compression workflow across diverse neural network architectures. Its design allows users to combine multiple compression techniques, including magnitude pruning, structured pruning, low-bit quantization, and quantization-aware training, within a single coherent API. Integration with Optuna enables reproducible and hardware-aware hyperparameter tuning, while MLflow supports experiment tracking and versioning. From a deployment perspective, the framework supports the transition from high-level model definitions to hardware-friendly compressed models, making it suitable for real-time and resource-constrained environments such as trigger systems. These features highlight PQuantML as both a research prototype and a practical engineering tool for building efficient neural networks.

Despite its flexibility, several opportunities for further development remain. In FITCompress, extended pre-training may lead to weight magnitudes growing excessively, increasing the minimum bit-width required for effective quantization beyond what is necessary. This could be prevented by constraining the weight range during the pre-training stage, for example through regularization or by applying high bit-width quantization. Another limitation concerns the compression metric, defined as the percentage of bit-operations (BOPs). While intuitive, this metric makes it difficult to estimate beforehand how many BOPs or EBOPs will remain after FITCompress, unless the baseline model complexity is known. A more practical alternative would be to allow users to specify a target EBOP budget directly. In addition, the current implementation of FITCompress is limited to PyTorch. The TensorFlow front-end of PQuantML also has several limitations, particularly in the layer replacement functionality. Currently, it supports only simple sequential architectures without branching, restricting its applicability to modern neural networks that incorporate residual connections, attention blocks, or multi-branch computational graphs. Furthermore, although PQuantML integrates with the hls4ml compiler, it currently lacks support for other hardware-specific compilation toolchains.

\section{Future Work}
\label{sec:future}
Future extensions of PQuantML will focus on addressing the limitations mentioned in the previous section, expanding backend support, improving automation, and incorporating additional model compression techniques. First, the full integration of the FITCompress method~\cite{zandonati2023optimalcompressionjointpruning}, when extended to TensorFlow, would unify the compression logic between backends and enable broader benchmarking of pruning and quantization strategies. Second, the hyperparameter optimization pipeline can be further simplified by introducing higher-level abstractions and automated search space generation. While the hyperparameter optimization loop efficiently manages sampling and pruning, many configuration steps remain manual. Therefore, automating the definition of parameter ranges, optimization strategies, and pruning schedules, potentially leveraging meta-learning from prior studies, would improve both usability and computational efficiency. Third, future work includes integrating a fully validated implementation of the FPGA-aware metric function~\cite{Ramhorst2023FPGA} within the MDMM method, and extending this approach to support more general constraint-based compression using diverse metric functions.

Finally, additional compression approaches, such as knowledge distillation~\cite{hinton2015distillingknowledgeneuralnetwork} and low-rank representation techniques~\cite{denil2013predicting}, offer promising directions for improving the trade-off between model accuracy and efficiency. For example, distillation can mitigate performance degradation caused by aggressive pruning or quantization, while low-rank factorization and tensor decomposition provide structured reductions in parameter count and computational cost. Integrating these techniques into PQuantML would enable more flexible and hardware-aware model optimization workflows.

\section{Conclusion}
\label{sec:conclusions}

In this work, we introduce PQuantML, a tool for training compressed deep learning models through a simple interface that integrates with hls4ml. The framework implements multiple pruning algorithms with different granularities and supports both layer-wise fixed-point quantization and High-Granularity Quantization (HGQ). The library enables the application of compression methods either directly via PQuantML layers or through an automated layer replacement function. 
We demonstrate the effectiveness of PQuantML on representative tasks such as jet substructure classification, comparing its performance with existing libraries such as QKeras and HGQ. Models trained with PQuantML using combined pruning and fixed-point quantization achieve strong performance while significantly reducing LUT and DSP usage as well as latency. In addition, PQuantML's HGQ-based models achieve performance comparable to models trained directly with HGQ, validating PQuantML as a reliable alternative to native HGQ training. Together with hls4ml, PQuantML provides a streamlined and flexible approach for training and deploying compressed models in hardware-constrained environments.

\section*{Acknowledgment}
This work has been funded by the Eric \& Wendy Schmidt Fund for Strategic Innovation through the CERN Next Generation Triggers project under grant agreement number SIF-2023-004. M.K. is supported by the US Department of Energy (DOE) under Grant No. DE-AC02-76SF00515.

\bibliographystyle{unsrt}
\bibliography{references}
\newpage
\appendix
\section{Appendix}
\subsection{Usage and structure}

\begin{lstlisting}[float=htbp,
    caption={Folder structure of PQuantML.},
    label={code:folder_structure},
    captionpos=b,
    style=folderstyle,
    emph={pquant,core,torch,keras,data_models,pruning_methods},
    emphstyle={\bfseries}
]
    pquant/
    ├── core/
    │   ├── torch/
    │   │   ├── activations
    │   │   ├── layers
    │   │   ├── fit_compress
    │   │   ├── quantizer
    │   │   └── train
    │   ├── keras/
    │   │   ├── activations
    │   │   ├── layers
    │   │   ├── quantizer
    │   │   └── train
    ├── data_models/ # Configuration objects
    ├── pruning_methods/ # Pruning methods
    ├── configs/
    ├── utils/
    └── examples/
\end{lstlisting}
\begin{figure}[h]
    \centering
    \begin{minipage}{0.92\linewidth}
        \begin{lstlisting}[language=Python,basicstyle=\small,style=pquant,label={lst:direct_model},caption={Model example using PQuantML layers directly.}]
import torch.nn as nn
from pquant.layers import PQConv2d, PQAvgPool2d, PQDense
from pquant.activations import PQActivation
from pquant import cs_config

class PQuantMLModel(nn.Module):
    def __init__(self, config):
        super().__init__()
        self.conv1 = PQConv2d(config, 3, 8, 3,
                              in_quant_bits=(1,0,7))
        self.relu1 = PQActivation(config, "relu")
        self.avg = PQAvgPool2d(config, 4)
        self.flatten = nn.Flatten()
        self.dense = PQDense(config, 72, 10,
                             quantize_output=True,
                             out_quant_bits=(1,2,5))

    def forward(self, x):
        # Forward pass

config = cs_config()
model = PQuantMLModel(config)
model(example_input)
\end{lstlisting}
    \end{minipage}
    \label{directlayer}
\end{figure}

\begin{figure}[ht]
    \centering
    \begin{minipage}{0.92\linewidth}
        \begin{lstlisting}[language=Python,basicstyle=\small,style=pquant,caption={Model example using automatic layer replacement.}, label={lst:layer_replacement}]
import torch.nn as nn
from pquant import add_compression_layers
from pquant import cs_config

class MLModel(nn.Module):
    def __init__(self):
        super().__init__()
        self.conv1 = nn.Conv2d(3, 8, 3)
        self.relu1 = nn.ReLU()
        self.avg = nn.AvgPool2d(4)
        self.flatten = nn.Flatten()
        self.dense = nn.Linear(72, 10)

    def forward(self, x):
        # Forward pass

model = MLModel()
config = cs_config()
model = add_compression_layers(model, config)
model(example_input)
\end{lstlisting}
    \end{minipage}
    \label{layerreplacement}

\end{figure}
\newpage
\begin{lstlisting}[language=Python,basicstyle=\tiny,  style=pquant, label=lst:train_model, caption=Usage of the generic training function.]
from pquant import cs_config, train_model

def train_epoch(model, epoch, training_data, loss_function **kwargs):
    # A training function
def validate_epoch(model, epoch, validation_data, loss_function, accuracy_function, **kwargs):
    # A validation function

train_model(model, config_cs, train_epoch, validate_epoch, training_data=training_data, validation_data=validation_data, loss_function=loss_function, accuracy_function=accuracy_function)
\end{lstlisting}

\subsection{Configuration parameters}
Tables showing the configurable training and quantization parameters in PQuantML. The training parameters configure the generic training function (Figure \ref{fig:training_loop}).
\begin{table}[h!]
    \centering
    \begin{tabular}{p{0.28\textwidth} p{0.65\textwidth}}
        \toprule
        \textbf{Parameter}   & \textbf{Explanation}                           \\
        \midrule
        pretraining\_epochs  &
        Number of epochs during the pretraining stage                         \\
        epochs               &
        Number of epochs during the main training stage                       \\
        fine\_tuning\_epochs &
        Number of epochs during the fine-tuning stage                         \\
        rounds               &
        Number of pruning/quantization rounds during training                 \\
        rewind               &
        When to rewind the weights:
        \textbf{never}, \textbf{every-round}, or \textbf{post-training-stage} \\
        save\_weights\_epoch &
        Epoch at which weights are saved for rewinding during the first round \\
        pruning\_first & Whether to prune before quantization. If false, pruning occurs after quantization \\
        \bottomrule
    \end{tabular}
    \caption{Training parameters defining the generic PQuantML training function.}
    \label{tab:training_parameters}
\end{table}

The quantization parameters configure how the model is quantized. It includes settings such as the default quantization bits for weights and data, whether to use HGQ or not and HGQ specific hyperparameters.

\begin{table}
    \centering
    \small
    \setlength{\tabcolsep}{6pt}
    \renewcommand{\arraystretch}{1.12}

    \begin{tabular}{p{0.33\textwidth} p{0.60\textwidth}}
        \toprule
        \textbf{Parameter}                & \textbf{Explanation}                                                              \\
        \midrule

        default\_data\_keep\_negatives    & Default \(k\) value for data quantization (0 or 1)                                \\

        default\_data\_integer\_bits      & Default integer bit-width \(i\) for data quantization                              \\

        default\_data\_fractional\_bits   & Default fractional bit-width \(f\) for data quantization                           \\

        default\_weight\_keep\_negatives  & Default \(k\) value for weight quantization (0 or 1)                              \\

        default\_weight\_integer\_bits    & Default integer bit-width \(i\) for weight quantization                            \\

        default\_weight\_fractional\_bits & Default fractional bitw-idth \(f\) for weight quantization                         \\

        granularity & Whether integer and fractional bits are the same for the whole weight matrix, shared between channels, or each weight has its own. Bit-widths stay the same for the whole tensor. \\

        quantize\_input                   & Whether inputs to layers are quantized by default                                 \\

        quantize\_output                  & Whether outputs of layers are quantized by default                                \\

        enable\_quantization              & Global switch to enable or disable quantization                                   \\
        hgq\_beta                         & HGQ loss coefficient scaling EBOPs                                                \\
        hgq\_gamma                        & HGQ loss coefficient regularizing bit-widths                                       \\
        layer\_specific                   & Dictionary containing quantization parameters for specific layers                 \\

        use\_high\_granularity\_quantization & Enable or disable HGQ                                                             \\

        use\_real\_tanh                   & Choose between real tanh and hard tanh                                            \\
        overflow\_mode\_data              & Overflow handling mode for data: SAT, SAT\_SYM, WRAP, WRAP\_SM                    \\
        overflow\_mode\_parameters        & Overflow handling mode for weights and biases: SAT, SAT\_SYM, WRAP, WRAP\_SM      \\

        round\_mode                       & Rounding mode: TRN, RND, RND\_CONV, RND\_ZERO, RN\_ZERO, RND\_MIN\_INF, RND\_INF. \\

        use\_relu\_multiplier             & Whether to use a learned bit-shift multiplier in ReLU inputs.                     \\

        \bottomrule
    \end{tabular}

    \caption{Quantization parameters of the configuration object.}
    \label{tab:quantization_parameters}
\end{table}

\begin{table}[h!]
    \centering
    \setlength{\tabcolsep}{3.15pt}
    \renewcommand{\arraystretch}{1.2}

    \begin{tabular}{lcccccc}
        \toprule
        \textbf{Model} & \textbf{Dataset} & \textbf{Pruning parameters} & \textbf{Epochs}  \\
        \midrule
        DST$^t$        & OpenML           & $\alpha: 1 \cdot 10^{-4}$                & 0/1500/500              \\
        DST$^w$        & OpenML           & $\alpha: 1.5 \cdot 10^{-4}$              & 0/200/100             \\
        PDP$^t$        & OpenML           & sparsity: 0.94              & 100/2000/200    &        \\
        PDP$^c$        & OpenML           & sparsity: 0.94              & 100/500/100     &        \\
        FITCompress$^c$         & OpenML           & compression goal: 0.0075    & 50/300/0        &        \\
        PDP$^t$        & CERN Box         & sparsity: 0.93              & 100/500/500     &      \\
        DST$^t$        & CERN Box         & $\alpha: 1 \cdot 10^{-4}$                  & 0/500/300       &       \\
        DST$^w$        & CERN Box         & $\alpha: 1.5 \cdot 10^{-4}$            & 0/200/100       &         \\
        DST$^t$        & PLF JSC          & $\alpha: 2.5 \cdot 10^{-5}$            & 0/600/500       &       \\
        \bottomrule
    \end{tabular}

    \caption{Hyperparameters used in the results shown in Table \ref{tab:resources_12816}. The learning rate is fixed to $1 \cdot 10^{-3}$ for all experiments.}
    \label{tab:training_hyperparams}

\end{table}

\end{document}